\documentclass[sigconf,nonacm]{acmart}

\setcopyright{none}
\renewcommand\footnotetextcopyrightpermission[1]{}
\AtBeginDocument{%
  }

\usepackage{algorithm}
\usepackage{algpseudocode}
\usepackage{amsthm}
\usepackage{multirow}
\usepackage{colortbl}         
\usepackage[most]{tcolorbox}  
\usepackage{pifont}           

\lstdefinestyle{sparql}{
  language=SPARQL,
  basicstyle=\ttfamily\small,
  escapeinside={(*@}{@*)},  
  keywordstyle=\color{blue!70!black},
  commentstyle=\itshape\color{gray!70!black},
  stringstyle=\color{purple!75!black},
  morekeywords={PREFIX, SELECT, DISTINCT, WHERE, FILTER, COUNT, GROUP, BY, ORDER, LIMIT, OPTIONAL, UNION, AS},
  frame=none,
  aboveskip=0pt,
  belowskip=0pt,
  showstringspaces=false,
}

\newtcolorbox{codeblock}{
  colback=gray!8,
  colframe=gray!8,
  boxrule=0pt,
  arc=2pt,
  left=8pt,
  right=8pt,
  top=4pt,
  bottom=4pt,
  boxsep=0pt,
}

\newtcolorbox{casestudy}[2][]{%
  title=Case #2: #1,
  colframe=blue!70!black,
  colback=gray!5,
  fonttitle=\bfseries,
  rounded corners,
  enhanced,
  breakable
}

\newcommand{\modelbox}[1]{\textbf{\textcolor{purple}{#1}}}
\newcommand{\sys}{\textsc{SchemaForge}}
\newcommand{\spsparql}{Spider4\allowbreak SPARQL}

\newcommand{\valstd}[2]{$#1_{\textcolor{blue}{\pm #2}}$}
\newcommand{\valstdb}[2]{$\mathbf{#1}_{\textcolor{blue}{\pm #2}}$}

\definecolor{rowgray}{gray}{0.94}
\definecolor{oursblue}{RGB}{222, 235, 252}

\begin{document}

\title{From Graph Retrieval to Schema Realization: Counterfactual Validation for Text-to-SPARQL over Heterogeneous\\ Knowledge Graphs}
\author{Chengxiao Dai, Yue Xiu and Dusit Niyato}

\begin{abstract}
Text-to-SPARQL maps natural-language questions to executable SPARQL queries over RDF knowledge graphs. While standard evaluations often fix the target graph in advance, practical knowledge graph question answering (KGQA) may involve heterogeneous graph collections with different schemas, partial alignments, and incomplete metadata. In this setting, query generation depends on more than SPARQL syntax: the system must identify a graph schema that can support the predicates, entity types, joins, filters, and constraints required by the question.

We present \sys{}, a schema-grounded agentic framework for text-to-SPARQL over
heterogeneous KG collections. Its central mechanism is question-conditioned schema-slice alignment: weak graph evidence first identifies plausible graphs, while stronger schema evidence determines whether a local schema slice can realize the intended query. The selected schema slice then constrains query generation and verification before execution. When only one graph is available, the same formulation reduces to standard single-KG text-to-SPARQL with schema grounding.

We evaluate \sys{} on LC-QuAD 2.0, QALD-9 Plus, QALD-10, and \spsparql{}. Across the four public benchmarks, \sys{} improves execution accuracy over the strongest matched agent baseline by 11.50 percentage points on average. On \spsparql{}, \sys{} improves execution accuracy from 54.86\% to 64.18\% and achieves 73.0\% Top-1 and 97.0\% Top-3 graph allocation accuracy. These results show that moving from weak graph evidence to schema-specific query commitments, together with counterfactual answer-set checks, improves executable query generation over heterogeneous knowledge graphs.
\end{abstract}






\maketitle

\section{Introduction}

Executable semantic parsing maps natural-language questions to inspectable
formal queries over structured data. In text-to-SPARQL, a question is answered
by generating a SPARQL query whose entities, predicates, filters, joins,
aggregation operators, and projected variables can be inspected before
execution. This explicit query structure is central to knowledge graph (KG)
question answering (KGQA), where correctness depends not only on the returned
answer but also on whether the query is grounded in the intended graph schema
~\citep{berant2013semantic,yih2015semantic,soru2017sparql,
banerjee2022modern,meyer2024assessing}.

Most text-to-SPARQL settings assume that the target graph is fixed before query
generation. Practical KGQA may instead involve heterogeneous graph collections
with different schemas, partial alignments, overlapping labels, and incomplete
metadata. This creates a graph-to-schema realization problem: before generating
SPARQL, a system must identify a graph schema that can express the predicates,
entity types, joins, filters, and constraints required by the question. Graph
descriptions, labels, aliases, and schema summaries can indicate topical
relevance, but they do not guarantee executable schema support. A graph may
mention the right concept while lacking the required predicate, and incomplete
summaries may omit domain--range constraints needed for valid triple patterns
~\citep{rahm2001survey,euzenat2013ontology,banerjee2022modern,
meyer2024assessing}.

Retrieval-augmented and graph-augmented generation methods provide external
evidence for knowledge-intensive tasks, but retrieved relevance alone does not
determine whether a graph schema licenses the triples and constraints used in a
SPARQL query~\citep{lewis2020rag,gao2023retrieval,edge2024local,
ji2024retrieval}. Federated SPARQL addresses a complementary problem: given an
already specified distributed query, federated engines optimize source
selection, join ordering, and adaptive execution~\citep{w3c-sparql,
sparql11-federated-query,schwarte2011fedx,acosta2011anapsid}. Our focus is the
preceding semantic parsing problem: generating executable SPARQL from natural
language when the compatible graph schema is not known in advance.

We introduce \sys{}, a schema-grounded agentic framework that moves from graph
retrieval to schema realization before query execution. The framework narrows
candidate graphs to local schema slices and uses these slices to constrain query
grounding, repair, and validation. Agentic inference is organized around typed
handoffs, including subgoal requirements, schema slices, structurally admissible
query candidates, and verified subgoal--graph--query records. This design ties
generation to graph-specific predicates, classes, and constraints rather than
retrieved context alone.

Schema realization constrains query construction, but it does not guarantee
query specificity. A query may be executable and schema-compatible while still
omitting a necessary type, predicate, join, or filter. To detect this failure
mode, \sys{} uses counterfactual answer-set validation: type-preserving
perturbations are applied to entity, predicate, and filter choices, and
candidates whose answer sets remain largely unchanged are revised or rejected
~\citep{yao2023react,wu2024autogen,guo2024large,he2024retrieving,
pan2023logic}.

We evaluate \sys{} on LC-QuAD~2.0, QALD-9 Plus, QALD-10, and \spsparql{}.
The first three benchmarks evaluate fixed-graph text-to-SPARQL, while
\spsparql{} evaluates graph allocation and query generation over heterogeneous
graph schemas~\citep{dubey2019lcquad2,perevalov2022qald9plus,
usbeck2024qald10,kosten2023spider4sparql}. Experiments show consistent gains
over matched agentic baselines, with the largest heterogeneous-graph
improvements on \spsparql{}.

\textbf{Contributions.}
This paper makes three contributions:
\begin{itemize}
    \item We formulate heterogeneous text-to-SPARQL as a graph-to-schema
    realization problem, where graph retrieval must be followed by
    schema-specific commitments before executable SPARQL can be generated.

    \item We propose \sys{}, a schema-grounded agentic framework in which
    specialized agents exchange typed artifacts, including subgoal requirements,
    schema slices, admissible query candidates, and verified
    subgoal--graph--query records.

    \item We introduce question-conditioned schema-slice alignment and
    counterfactual answer-set validation to constrain query generation and
    detect executable but underspecified SPARQL.
\end{itemize}

\section{Related Work}

\textbf{Executable KGQA, text-to-SPARQL, and schema selection.}
Text-to-SPARQL treats KGQA as executable semantic parsing, where a
natural-language question is mapped to a SPARQL query over an RDF knowledge
graph. Prior work has studied staged query-graph generation, template-based
querying, embedding-based multi-hop reasoning, graph-neural reasoning, and
prompt-based KG reasoning
~\citep{yih2015semantic,saxena2020improving,yasunaga2021qa,cui2024prompt,
10.1007/978-3-031-72344-5_17}. These methods address entity linking, relation
grounding, and compositional reasoning, but typically assume that the target
graph or schema is fixed before query generation. Schema matching, ontology
matching, and data integration study correspondences among classes, properties,
entities, and constraints across heterogeneous sources
~\citep{rahm2001survey,euzenat2013ontology,dong2013data}. In heterogeneous
text-to-SPARQL, however, graph selection is question-conditioned: a source is
useful only if its schema can support the query intent. Our setting therefore
requires schema selection as part of executable query generation.

\textbf{Retrieval, graph-augmented generation, and agentic reasoning.}
Retrieval-augmented and graph-augmented methods retrieve passages, entities,
paths, neighborhoods, graph summaries, or modular evidence chains for
knowledge-intensive reasoning~\citep{xu2024search,han2024retrieval,
wu2025composerag,wang2024astute,ji2024retrieval}. Such evidence can identify
relevant context, but text-to-SPARQL also requires schema support for the
triples, joins, and constraints used in the generated query. Agentic frameworks
coordinate language-model calls or agents for planning, tool use, role-based
collaboration, and multi-step problem solving
~\citep{li2023camel,hong2023metagpt,wu2024autogen,tran2025multi,
barbosa2024collaborative,guo2024large}. \sys{} follows this agentic view while
constraining handoffs with graph, schema, and execution states rather than
unconstrained natural-language exchanges.

\textbf{Verification and distributed SPARQL execution.}
Verification-augmented generation uses external checks, self-revision, symbolic
constraints, or tool feedback to improve generated outputs
~\citep{he2024retrieving,pan2023logic,vakharia2024proslm,press2022measuring,
ji2023survey}. Grammar- and shape-constrained decoding approaches such as
PICARD for SQL~\citep{scholak2021picard} and SHACL-based RDF validation~\citep{w3c-shacl}
enforce syntactic or structural validity at generation or post-hoc,
but neither addresses underspecified queries that are valid yet miss
discriminative slots. For text-to-SPARQL, execution success alone is insufficient,
because an underspecified query may still parse and return answers. Our work
uses structural validation and counterfactual answer-set tests to target both
invalid SPARQL and executable but underspecified queries. Federated SPARQL is
complementary: SPARQL~1.1 defines query and federated-query semantics, including
distributed access through \texttt{SERVICE}
~\citep{w3c-sparql,sparql11-federated-query,prudhommeaux2013sparql}, while
systems such as FedX and ANAPSID optimize source selection, join ordering, and
adaptive execution for already specified distributed queries
~\citep{schwarte2011fedx,acosta2011anapsid}. Heterogeneous text-to-SPARQL
addresses the preceding problem of selecting a compatible graph schema and
generating graph-scoped SPARQL from natural language.

\begin{figure*}[!ht]
    \centering
    \includegraphics[width=\linewidth]{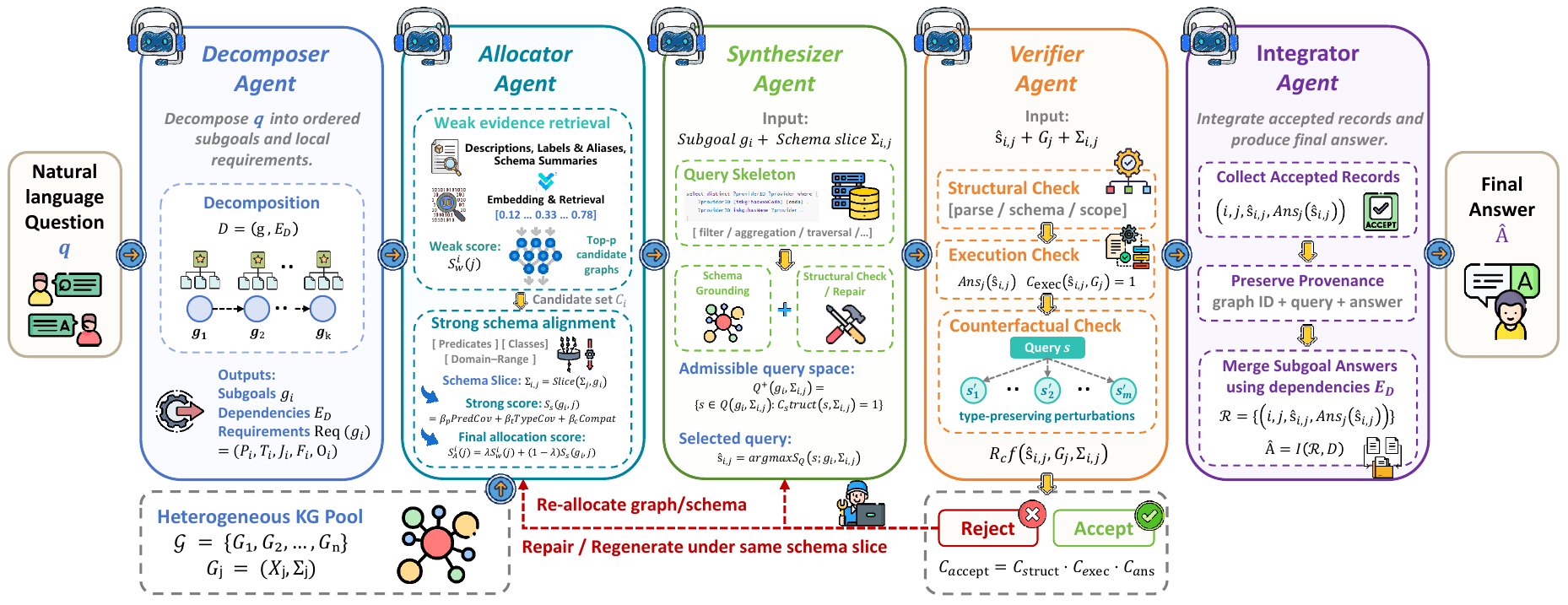}
    \caption{Overview of the \sys{} framework.}
    \label{fig:workflow}
\end{figure*}

\section{Problem Formulation}
\label{sec:problem}

We consider text-to-SPARQL over a collection of heterogeneous RDF graphs. Let
\(\mathcal{G}=\{G_1,\ldots,G_n\}\) denote the graph collection and
\([n]=\{1,\ldots,n\}\). Each graph \(G_j\) has a triple set
\[
X_j=\{(u,p,v):(u,p,v)\in G_j\}
\]
and an associated graph summary \(\Sigma_j\). The summary contains the schema
and graph metadata used for grounding, including classes, predicates, labels,
aliases, namespace declarations, named-graph identifiers, domain--range
constraints, and entity metadata when available.

Given a natural-language question \(q\), the task is to generate executable
SPARQL queries over one or more graphs in \(\mathcal{G}\) and return a final
answer \(\hat{A}\). A graph-scoped query is evaluated either against the
endpoint associated with \(G_j\) or against the named graph corresponding to
\(G_j\) in a shared endpoint. Unlike fixed-endpoint text-to-SPARQL, the relevant
graph indices are not assumed to be known in advance.

A prediction is represented as
\[
y=(D,\mathbf{a},\hat{\mathbf{s}},\hat{A}).
\]
The decomposition is \(D=(\mathbf{g},E_D)\), where
\(\mathbf{g}=(g_1,\ldots,g_k)\) is the ordered subgoal sequence,
\([k]=\{1,\ldots,k\}\), and \(E_D\subseteq[k]\times[k]\) is a set of directed
dependency edges. An edge \((r,i)\in E_D\) indicates that subgoal \(g_i\)
depends on the output of subgoal \(g_r\). We assume that \(E_D\) is acyclic and
that \(\mathbf{g}\) is topologically ordered, so \((r,i)\in E_D\) implies
\(r<i\). If no explicit dependency is required, \(E_D=\emptyset\).

The vector \(\mathbf{a}=(a_1,\ldots,a_k)\) is the final verified graph
assignment, where \(a_i\subseteq[n]\) contains the graph indices accepted for
subgoal \(g_i\). During inference, the system may maintain a retrieval set
\(C_i\subseteq[n]\) and a provisional assignment \(\tilde{a}_i\subseteq C_i\).
Only accepted subgoal--graph records associated with \(a_i\) are used for
answer integration.

The accepted query family is indexed by subgoal and graph:
\[
\hat{\mathbf{s}}
=
\{(i,j,\hat{s}_{i,j}): i\in[k],\ j\in a_i\},
\]
where \(\hat{s}_{i,j}\) is the selected SPARQL query for subgoal \(g_i\) on
graph \(G_j\). For a query \(s\) executed on \(G_j\), \(\mathrm{Ans}_j(s)\)
denotes the canonical set of projected solution mappings returned by \(s\),
after duplicate elimination. If execution fails, \(\mathrm{Ans}_j(s)\) is
undefined.

For each subgoal--graph pair \((g_i,G_j)\), let
\[
\Sigma_{i,j}=\operatorname{Slice}(\Sigma_j,g_i)
\]
be the summary slice selected for \(g_i\) from \(G_j\). The operator
\(\operatorname{Slice}\) returns the schema and metadata elements relevant to
grounding and validating the subgoal.

A candidate query \(s\) is structurally admissible for \((g_i,G_j)\) if it
satisfies three binary checks. First, it must parse under the SPARQL grammar:
\[
C_{\mathrm{parse}}(s)=1.
\]
Second, IRIs introduced during grounding must be contained in or mapped by the
selected summary slice:
\[
C_{\mathrm{schema}}(s,\Sigma_{i,j})=1.
\]
This check covers predicates, classes used in type constraints, named-graph
identifiers, and grounded entity IRIs, and also checks prefix resolution. Third,
variables must be valid under SPARQL scoping rules:
\[
C_{\mathrm{scope}}(s)=1.
\]
Directly projected variables must be bound in the query pattern, while variables
introduced by expressions of the form \((E\ \mathrm{AS}\ ?v)\), where \(E\) is a
SPARQL expression, are valid projected aliases. Variables used in filters,
grouping, ordering, and aggregate expressions must be bound in the appropriate
scope. In aggregate queries, projected variables must be grouping variables,
aggregate expressions, or valid expression aliases.

The combined structural admissibility indicator is
\[
C_{\mathrm{struct}}(s,\Sigma_{i,j})
=
C_{\mathrm{parse}}(s)\,
C_{\mathrm{schema}}(s,\Sigma_{i,j})\,
C_{\mathrm{scope}}(s),
\]
where each factor is in \(\{0,1\}\). Structural admissibility captures syntactic
and schema-level validity, but it does not ensure that the query answers the
intended subgoal. A structurally valid query may still omit a required predicate,
type, join, or filter. Generated queries are therefore verified before answer
integration, as described in Section~\ref{sec:maf}. When \(n=1\), graph
selection reduces to the single-graph case, leaving schema-grounded
text-to-SPARQL with verification.

\section{System Design}
\label{sec:maf}

Figure~\ref{fig:workflow} illustrates \sys{} as a schema-grounded agentic
framework. The framework contains five specialized agents: a
\emph{Decomposer}, an \emph{Allocator}, a \emph{Synthesizer}, a
\emph{Verifier}, and an \emph{Integrator}. Algorithm~\ref{alg:agentic_t2s}
summarizes their joint inference procedure; notation is listed in
Appendix~\ref{app:notation}.

Inference instantiates the prediction \(y\) by constructing accepted
subgoal--graph--query records. For each subgoal \(g_i\), the graph search space
is restricted from \([n]\) to a retrieval set \(C_i\), then to a provisional
assignment \(\tilde a_i\) using schema evidence. For each
\(j\in\tilde a_i\), query generation is restricted to structurally admissible
candidates in \(\mathcal{Q}^{+}(g_i,\Sigma_{i,j})\). Verification converts
\(\tilde a_i\) into the final assignment \(a_i\), and the accepted records are
integrated into \(\hat A\).

The same intermediate objects are used across stages:
\(\operatorname{Req}(g_i)\) guides graph allocation, query grounding, and
empty-answer verification, while \(\Sigma_{i,j}\) constrains both candidate
generation and structural validation.


\begin{algorithm}[!ht]
\caption{\sys{}: Multi-Agent Text-to-SPARQL Inference}
\label{alg:agentic_t2s}
\begin{algorithmic}[1]
\Require Question $q$; graph pool $\mathcal{G}=\{(G_j,\Sigma_j)\}_{j=1}^{n}$;
         budgets $(p,r,m,\rho)$, thresholds $(\tau_A,\tau_{\mathrm{cf}})$, weight $\lambda$
\Ensure Verified answer $\hat A$
\State $D=(\mathbf{g},E_D)\gets\textsc{Decomposer}(q)$
  \Comment{$k$ subgoals + dependencies}
\State $\mathcal{R}\gets\emptyset$
\For{each subgoal $g_i$ in a topological order of $E_D$}
  \State $H_i\gets\textsc{ResolveDeps}(\mathcal{R},D,i)$
    \Comment{dependency context for $g_i$}
  \State $\tilde a_i\gets\textsc{Allocator}(g_i,H_i,\mathcal{G})$
    \Comment{weak retrieval $\to$ schema-aware rerank}
  \State $a_i\gets\emptyset$
  \For{each $j\in\tilde a_i$}
    \State $\Sigma_{i,j}\gets\textsc{Slice}(\Sigma_j,g_i)$
    \State $\hat s_{i,j}\gets\textsc{Synthesizer}(g_i,H_i,\Sigma_{i,j})$
      \Comment{template grounding + repair}
    \If{$\hat s_{i,j}\neq\bot$}
      \State $(v_{i,j},B_{i,j})\gets
        \textsc{Verifier}(\hat s_{i,j},G_j,\Sigma_{i,j},g_i,H_i)$
      \If{$v_{i,j}=\mathsf{accept}$}
        \State $a_i\gets a_i\cup\{j\}$
        \State $\mathcal{R}\gets\mathcal{R}\cup
          \{(i,j,g_i,G_j,\hat s_{i,j},B_{i,j})\}$
      \EndIf
    \EndIf
  \EndFor
\EndFor
\State $\hat A\gets\textsc{Integrator}(\mathcal{R},D,\mathcal{G})$
\State \Return $\hat A$
  \Comment{each $(i,j,\hat s_{i,j},B_{i,j})\!\in\!\mathcal{R}$ is structurally admissible, executes on $G_j$, and satisfies $C_{\mathrm{ans}}$}
\end{algorithmic}
\end{algorithm}

\subsection{Subgoal Interpretation (Decomposer Agent)}

\sys{} uses the decomposition \(D=(\mathbf{g},E_D)\). Each subgoal \(g_i\)
corresponds to an operation such as entity lookup, filtering, comparison,
aggregation, or relation traversal, and may include entity mentions, expected
answer types, and explicit constraints.

We represent the inferred local grounding requirements of \(g_i\) as
\[
\operatorname{Req}(g_i)
=
(\mathcal{P}_i,\mathcal{T}_i,\mathcal{J}_i,\mathcal{F}_i,\mathcal{O}_i),
\]
where \(\mathcal{P}_i\) denotes predicate or relation requirements,
\(\mathcal{T}_i\) denotes type requirements, \(\mathcal{J}_i\) denotes join
requirements, \(\mathcal{F}_i\) denotes filters or constraints, and
\(\mathcal{O}_i\) denotes the expected output structure.

\subsection{Weak-to-Strong Graph Allocation (Allocator Agent)}

For subgoal \(g_i\), allocation first constructs a retrieval set \(C_i\) using
textual similarity. Let \(d_j\) be a textual serialization of graph \(G_j\),
including its description, labels, aliases, and graph summary. The weak
retrieval score is
\[
S_w^i(j)
=
\begin{cases}
\dfrac{1+\cos(e(g_i),e(d_j))}{2},
& e(g_i)\neq 0 \ \text{and}\ e(d_j)\neq 0,\\[6pt]
0,
& \text{otherwise},
\end{cases}
\]
where \(e(\cdot)\) is the embedding function.

Let \(\operatorname{Top}_b(U,S)\) return up to \(b\) elements of \(U\) with the
largest values under score function \(S\), using deterministic tie breaking. The
retrieval set is
\[
C_i=\operatorname{Top}_p([n],S_w^i),
\]
where \(|C_i|\le \min(p,n)\).

The strong allocation stage reranks \(C_i\) using schema compatibility. For each
\(j\in C_i\), it extracts
\[
\Sigma_{i,j}=\operatorname{Slice}(\Sigma_j,g_i)
\]
and computes
\[
\begin{aligned}
S_s(g_i,j)
={}& \beta_p\,\mathrm{PredCov}(g_i,\Sigma_{i,j}) \\
  & + \beta_t\,\mathrm{TypeCov}(g_i,\Sigma_{i,j}) \\
  & + \beta_c\,\mathrm{Compat}(g_i,\Sigma_{i,j}).
\end{aligned}
\]
Here \(\mathrm{PredCov}\) measures predicate or relation coverage,
\(\mathrm{TypeCov}\) measures expected type coverage, and
\(\mathrm{Compat}\) measures compatibility between candidate
predicate--argument bindings and available domain--range constraints. Missing
domain--range constraints are treated as unobserved rather than contradictory.
The component scores and weights satisfy
\[
\begin{aligned}
\mathrm{PredCov},\ \mathrm{TypeCov},\ \mathrm{Compat} &\in [0,1], \\
\beta_p,\ \beta_t,\ \beta_c &\ge 0, \\
\beta_p + \beta_t + \beta_c &= 1.
\end{aligned}
\]

The final allocation score is
\[
S_A^i(j)
=
\lambda S_w^i(j)
+
(1-\lambda)S_s(g_i,j),
\]
where \(\lambda\in[0,1]\). The provisional assignment is
\[
T_i=\{j\in C_i:S_A^i(j)\ge \tau_A\},
\qquad
\tilde{a}_i=\operatorname{Top}_r(T_i,S_A^i),
\]
where \(r\) is the strong-stage budget and \(\tau_A\in[0,1]\) is the allocation
threshold.

\subsection{Schema-Grounded SPARQL Generation (Synthesizer Agent)}

The provisional assignment \(\tilde a_i\) defines the subgoal--graph pairs for
which query generation is attempted. For each \(j\in\tilde a_i\), \sys{}
constructs a schema-constrained candidate space
\(\mathcal{Q}(g_i,\Sigma_{i,j})\). Candidates are generated by grounding query
templates in the selected summary slice. A template specifies an abstract SPARQL
form, such as lookup, filtering, comparison, aggregation, or relation traversal.

Grounding instantiates schema-dependent slots with predicates, classes, entity
IRIs, and named-graph identifiers contained in or mapped by \(\Sigma_{i,j}\).
Literal constraints are derived from \(q\) and checked for compatibility with
the selected summary slice. The skeleton library is in
Appendix~\ref{app:sparql_templates}.

Only structurally admissible candidates are retained:
\[
\mathcal{Q}^{+}(g_i,\Sigma_{i,j})
=
\{s\in\mathcal{Q}(g_i,\Sigma_{i,j}):
C_{\mathrm{struct}}(s,\Sigma_{i,j})=1\}.
\]
If \(\mathcal{Q}^{+}(g_i,\Sigma_{i,j})=\emptyset\), generation fails for the
pair. Otherwise, the selected query is
\[
\hat{s}_{i,j}
=
\operatorname*{arg\,max}_{s\in\mathcal{Q}^{+}(g_i,\Sigma_{i,j})}
S_Q(s;g_i,\Sigma_{i,j}),
\]
with deterministic tie breaking. The score \(S_Q\) measures compatibility
between the candidate query, the subgoal, and the selected summary slice.

Before rejecting a candidate, a deterministic repair rule checks prefix
resolution, predicate and class availability, entity grounding, filter
compatibility, projection aliases, SPARQL scoping, and schema compatibility. If
a unique schema-compatible correction exists under this rule, it is applied.
Otherwise, the candidate is regenerated under the same summary slice until the
retry budget \(\rho\) is exhausted. If generation or repair fails,
\(\hat{s}_{i,j}\) is undefined and the pair is not accepted.

\subsection{Verification (Verifier Agent)}

Verification decides whether each generated query \(\hat{s}_{i,j}\) is accepted.
A candidate \(s\) must be structurally admissible and executable, where
\[
C_{\mathrm{exec}}(s,G_j)=1
\quad\Longleftrightarrow\quad
\mathrm{Ans}_j(s)\ \text{is defined}.
\]
For candidates that pass these checks, let \(B=\mathrm{Ans}_j(s)\).

For non-empty answers, the verifier tests whether the answer changes under
type-preserving perturbations. Let \(\operatorname{proj}(s)\) be the ordered
projected output signature of \(s\). The valid perturbation set is
\[
\mathcal{V}_j(s,\Sigma_{i,j})
=
\left\{
s'=T(s):
\begin{array}{l}
T\in\mathcal{T}(s,\Sigma_{i,j}),\quad s'\neq s,\\
C_{\mathrm{struct}}(s',\Sigma_{i,j})=1,\\
C_{\mathrm{exec}}(s',G_j)=1,\\
\operatorname{proj}(s')=\operatorname{proj}(s)
\end{array}
\right\}.
\]
The verifier uses a subset
\(\mathcal{V}^{(m)}_j(s,\Sigma_{i,j})\subseteq\mathcal{V}_j(s,\Sigma_{i,j})\)
with
\[
|\mathcal{V}^{(m)}_j(s,\Sigma_{i,j})|
=
\min(m,|\mathcal{V}_j(s,\Sigma_{i,j})|).
\]
When \(B\neq\emptyset\) and
\(|\mathcal{V}^{(m)}_j(s,\Sigma_{i,j})|>0\), the counterfactual invariance score
is
\[
R_{\mathrm{cf}}(s,G_j,\Sigma_{i,j})
=
\frac{1}{|\mathcal{V}^{(m)}_j(s,\Sigma_{i,j})|}
\sum_{s'\in\mathcal{V}^{(m)}_j(s,\Sigma_{i,j})}
\frac{|B\cap \mathrm{Ans}_j(s')|}
{|B\cup \mathrm{Ans}_j(s')|}.
\]
Large \(R_{\mathrm{cf}}\) indicates that the answer is insensitive to changes in
entity, predicate, or filter choices.

Empty answers are not scored by counterfactual invariance. Instead, they are
accepted only if the query covers the required subgoal components:
\[
C_{\mathrm{empty}}(s,\Sigma_{i,j},g_i)=1.
\]
Concretely, \(C_{\mathrm{empty}}=1\) iff every predicate and type required by \(\operatorname{Req}(g_i)\) appears in the WHERE clause of \(s\) and is resolved in \(\Sigma_{i,j}\); this ensures that an empty result reflects genuine data absence rather than schema-coverage failure.

The answer-dependent acceptance condition is
\[
C_{\mathrm{ans}}(s,G_j,\Sigma_{i,j},g_i)
=
\begin{cases}
1, & \text{if } B\neq\emptyset \text{ and} \\
   & \quad R_{\mathrm{cf}}\le\tau_{\mathrm{cf}}, \\[2pt]
1, & \text{if } B=\emptyset \text{ and} \\
   & \quad C_{\mathrm{empty}}=1, \\[2pt]
0, & \text{otherwise.}
\end{cases}
\]
The final acceptance indicator is
\[
C_{\mathrm{accept}}(s,G_j,\Sigma_{i,j},g_i)
=
C_{\mathrm{struct}}(s,\Sigma_{i,j})\,
C_{\mathrm{exec}}(s,G_j)\,
C_{\mathrm{ans}}(s,G_j,\Sigma_{i,j},g_i).
\]

The final verified graph assignment for \(g_i\) is
\[
a_i
=
\left\{
j\in\tilde{a}_i:
\hat{s}_{i,j}\ \text{is defined and}\
C_{\mathrm{accept}}(\hat{s}_{i,j},G_j,\Sigma_{i,j},g_i)=1
\right\}.
\]
Only records indexed by \(a_i\) are passed to answer integration.

\subsection{Answer Integration (Integrator Agent)}

The accepted assignment defines the records consumed by the integrator:
\[
\mathcal{R}
=
\left\{
(i,j,g_i,G_j,\hat{s}_{i,j},\mathrm{Ans}_j(\hat{s}_{i,j})):
i\in[k],\ j\in a_i
\right\}.
\]
The final answer is
\[
\hat{A}=I(\mathcal{R},D).
\]
The integration function normalizes identifiers and literals, removes
duplicates, aligns equivalent entities when metadata are available, and
preserves graph and query provenance. For compositional questions, it uses the
dependency structure \(E_D\), including joins, intersections, comparisons,
ranking, and aggregations over intermediate answer sets.

\subsection{Hyperparameters and Execution Invariant}

The thresholds \(\tau_A\) and \(\tau_{\mathrm{cf}}\), weights
\(\lambda,\beta_p,\beta_t,\beta_c\), weak-retrieval budget \(p\),
strong-stage budget \(r\), perturbation budget \(m\), and retry budget \(\rho\)
are selected on validation data, with
\[
\tau_A,\tau_{\mathrm{cf}},\lambda\in[0,1],
\qquad
p,r,m,\rho\in\mathbb{N}_{+}.
\]

Every record passed to the integrator satisfies structural admissibility,
successful execution, and one acceptance condition. Thus each integrated record
is executable and verifier-admissible. Experiments report execution accuracy,
triple-level F1, graph-allocation accuracy, and ablations.

\section{Experiments}
\label{sec:experiments}

\subsection{Experimental Setup and Benchmarks}
\label{subsec:setup}

We evaluate \sys{} on text-to-SPARQL in both fixed-graph and heterogeneous-graph settings. Predictions are scored by executing the generated SPARQL against the target graph and comparing the answer set with gold. The Wikidata benchmarks fix the target endpoint; \spsparql{} requires selecting one of 20 candidate graphs from schema summaries before generation. All matched-backbone methods share the same backbone large language model (LLM, gpt-oss-120b), decoding settings, schema-context token budget, and repair budget $\rho$; all backbone LLMs are accessed through their official APIs or local deployments. We report mean and standard deviation over three random seeds; full configuration is in Appendix~\ref{subsec:config}.


\begin{table}[!ht]
\centering
\caption{Evaluation benchmarks.}
\label{tab:benchmarks}
\small
\setlength{\tabcolsep}{5pt}
\begin{tabular}{lccc}
\toprule
\textbf{Benchmark} & \textbf{Role} & \textbf{Sources} & \textbf{Multi-source?}\\
\midrule
LC-QuAD~2.0 & single-source parsing & 1 & \textcolor{red}{\ding{55}}\\
QALD-9 Plus & single-source parsing & 1 & \textcolor{red}{\ding{55}}\\
QALD-10 & single-source parsing & 1 & \textcolor{red}{\ding{55}}\\
\spsparql{} & large-pool routing & 20 & \textcolor{blue}{\ding{52}}\\
\bottomrule
\end{tabular}
\end{table}

We use four public benchmarks. LC-QuAD~2.0, QALD-9 Plus, and QALD-10 evaluate text-to-SPARQL over a fixed Wikidata endpoint. \spsparql{} evaluates query generation over a heterogeneous graph pool and provides gold source annotations for graph-allocation analysis~\citep{dubey2019lcquad2,perevalov2022qald9plus,usbeck2024qald10,kosten2023spider4sparql}. The main aggregate result is an unweighted macro-average over these four public benchmarks.

\subsection{Baselines and Metrics}
\label{subsec:baselines_metrics}

\textbf{Baselines.}
We compare \sys{} with matched-backbone baselines. \textbf{Vanilla LLM} directly generates SPARQL from the question and schema context. \textbf{ReAct}~\citep{yao2023react} and \textbf{AutoGen}~\citep{wu2024autogen} use the same schema context and SPARQL execution tools as \sys{} but do not use weak-to-strong graph allocation, schema-grounded skeleton generation, or the dual-stage verifier. On \spsparql{}, we additionally compare against \textbf{CoT-Planner Agent}~\citep{wei2022chain}, a chain-of-thought planning baseline that decomposes the question and drafts a SPARQL skeleton before slot filling, and evaluate graph allocation against a zero-shot LLM routing baseline. We also report external fine-tuned systems as reference when available: \textbf{MST5}~\citep{srivastava2024mst5}, \textbf{Spinach}~\citep{liu2024spinach}, and \textbf{ValueNet4SPARQL}~\citep{kosten2023spider4sparql}; these are excluded from the matched-backbone macro average.

\textbf{Metrics.} We report \textbf{Execution Accuracy (EA)}, the fraction of queries whose answer sets exactly match the gold set; \textbf{Query Syntax Correctness (QSC)}, the fraction that compile under Apache Jena; \textbf{Triple-Level F1 (TF1)}, micro-averaged F1 over WHERE-clause triple patterns ignoring variable renaming and \texttt{FILTER}s; and \textbf{Macro F1 (MF1)}, per-question F1 between predicted and gold answer sets averaged across questions (QALD protocol). For \spsparql{}, we additionally report \textbf{Top-1} and \textbf{Top-3} graph-allocation accuracy. To compare inference cost across methods, we report \textbf{Average Token Usage (ATU)}, the mean total (input + output) token count per prediction across all LLM calls. Formal definitions are in Appendix~\ref{app:metrics}.

\begin{table*}[!ht]
\centering
\caption{Main results on the three Wikidata benchmarks.}
\label{tab:wikidata_results}
\setlength{\tabcolsep}{18pt}
\renewcommand{\arraystretch}{1.05}
\begin{tabular}{llcccc}
\toprule
\textbf{Dataset} & \textbf{Method} & \textbf{EA (\%)} & \textbf{QSC (\%)} & \textbf{TF1 (\%)} & \textbf{MF1 (\%)} \\
\midrule

& MST5 (fine-tuned)
  & \valstd{42.35}{1.12} & \valstd{56.48}{1.36} & \valstd{48.92}{0.94} & \valstd{46.17}{1.28} \\
\rowcolor{rowgray}
\cellcolor{white} & Vanilla LLM
  & \valstd{37.84}{1.47} & \valstd{52.93}{1.21} & \valstd{43.71}{1.58} & \valstd{40.58}{1.33} \\
& Spinach
  & \valstd{54.26}{0.88} & \valstd{67.08}{1.42} & \valstd{59.13}{1.07} & \valstd{57.44}{1.31} \\
\rowcolor{rowgray}
\cellcolor{white} & ReAct
  & \valstd{48.91}{2.06} & \valstd{64.73}{1.61} & \valstd{54.85}{1.84} & \valstd{52.06}{1.72} \\
& AutoGen
  & \valstd{52.47}{1.74} & \valstd{66.29}{2.18} & \valstd{58.64}{1.46} & \valstd{55.73}{1.93} \\
\rowcolor{oursblue}
\cellcolor{white}\multirow{-6}{*}{\textit{LC-QuAD 2.0}} & \textbf{\textsc{SchemaForge} (Ours)}
  & \valstdb{63.18}{1.21} & \valstdb{74.96}{1.54} & \valstdb{68.29}{1.18} & \valstdb{66.41}{1.43} \\
\midrule

& MST5 (fine-tuned)
  & \valstd{51.76}{1.05} & \valstd{65.28}{1.44} & \valstd{57.31}{1.16} & \valstd{55.84}{1.37} \\
\rowcolor{rowgray}
\cellcolor{white} & Vanilla LLM
  & \valstd{44.39}{1.62} & \valstd{59.12}{1.29} & \valstd{49.83}{1.51} & \valstd{48.26}{1.18} \\
& Spinach
  & \valstd{65.12}{0.91} & \valstd{78.46}{1.33} & \valstd{69.38}{1.48} & \valstd{70.21}{1.12} \\
\rowcolor{rowgray}
\cellcolor{white} & ReAct
  & \valstd{55.74}{1.87} & \valstd{70.05}{1.56} & \valstd{61.19}{2.14} & \valstd{60.04}{1.69} \\
& AutoGen
  & \valstd{59.28}{1.53} & \valstd{72.81}{2.03} & \valstd{64.33}{1.72} & \valstd{63.27}{1.86} \\
\rowcolor{oursblue}
\cellcolor{white}\multirow{-6}{*}{\textit{QALD-9 Plus}} & \textbf{\textsc{SchemaForge} (Ours)}
  & \valstdb{72.36}{1.24} & \valstdb{83.42}{1.46} & \valstdb{77.31}{1.28} & \valstdb{76.94}{1.39} \\
\midrule

& MST5 (zero-shot transfer)
  & \valstd{32.46}{1.34} & \valstd{45.18}{1.51} & \valstd{37.42}{1.27} & \valstd{36.08}{1.43} \\
\rowcolor{rowgray}
\cellcolor{white} & Vanilla LLM
  & \valstd{29.74}{1.73} & \valstd{42.58}{1.46} & \valstd{35.11}{1.64} & \valstd{33.86}{1.28} \\
& Spinach
  & \valstd{48.95}{1.16} & \valstd{61.37}{1.49} & \valstd{53.76}{1.21} & \valstd{54.68}{1.54} \\
\rowcolor{rowgray}
\cellcolor{white} & ReAct
  & \valstd{40.36}{2.21} & \valstd{53.95}{1.68} & \valstd{45.72}{1.97} & \valstd{44.15}{2.08} \\
& AutoGen
  & \valstd{43.58}{1.69} & \valstd{56.11}{2.26} & \valstd{49.37}{1.74} & \valstd{47.96}{1.91} \\
\rowcolor{oursblue}
\cellcolor{white}\multirow{-6}{*}{\textit{QALD-10}} & \textbf{\textsc{SchemaForge} (Ours)}
  & \valstdb{56.48}{1.35} & \valstdb{68.21}{1.61} & \valstdb{61.63}{1.30} & \valstdb{60.72}{1.51} \\
\bottomrule
\end{tabular}

\end{table*}

\subsection{Main Results}
\label{subsec:main_results}

Tables~\ref{tab:wikidata_results} and~\ref{tab:spider_results} report main public-benchmark results across the fixed-graph (Wikidata) and heterogeneous (\spsparql{}) settings.

\textbf{Fixed-graph Wikidata benchmarks.}
On LC-QuAD~2.0, QALD-9 Plus, and QALD-10, \sys{} consistently improves EA, QSC, TF1, and MF1 over all matched-backbone baselines. Compared with AutoGen, \sys{} improves EA by $+10.71$ points on LC-QuAD~2.0, $+13.08$ points on QALD-9 Plus, and $+12.90$ points on QALD-10. The gains are stable across benchmarks despite considerable variation in baseline strength (AutoGen ranges from $43.58\%$ on QALD-10 to $59.28\%$ on QALD-9 Plus), indicating that schema-grounded query construction and verification contribute beyond what stronger prompting and retry-based agentic execution provide. Compared with the strongest agentic reference baseline, Spinach, \sys{} improves EA by $+8.92$ on LC-QuAD~2.0, $+7.24$ on QALD-9 Plus, and $+7.53$ on QALD-10; against the fine-tuned MST5 baseline, the gains are larger ($+20.83$, $+20.60$, and $+24.02$ respectively), consistent with the hypothesis that schema grounding and counterfactual verification provide compounding returns when paired with strong general-purpose LLMs.

\textbf{Heterogeneous graph benchmark.}
\sys{} improves EA from $54.86\%$ to $64.18\%$ over AutoGen ($+9.32$ EA) on \spsparql{}, with gains of $+7.91$ in QSC and $+9.36$ in TF1. The relatively smaller gap to AutoGen on \spsparql{}, compared with Wikidata, is consistent with AutoGen's iterative tool-use compensating for some allocation errors at the cost of more SPARQL execution calls; the routing analysis in \S\ref{subsec:graph_allocation} isolates the allocation component, where \sys{} improves Top-1 routing from $65.0\%$ to $73.0\%$.

\textbf{Aggregate performance.}
Using an unweighted macro-average over the four public benchmarks, \sys{} reaches $64.05\%$ EA versus $52.55\%$ for AutoGen, a $+11.50$ EA absolute improvement. The per-benchmark gains over AutoGen are consistent (\spsparql{} $+9.32$; LC-QuAD~2.0 $+10.71$; QALD-9 Plus $+13.08$; QALD-10 $+12.90$), indicating that the contribution is not concentrated in any single benchmark family.

\begin{table*}[t]
  \centering
  \caption{Main results on \spsparql{}.}
  \label{tab:spider_results}
  \setlength{\tabcolsep}{14pt}
  \renewcommand{\arraystretch}{1.05}
  \begin{tabular}{llcccc}
  \toprule
  \textbf{Dataset} & \textbf{Method} & \textbf{EA (\%)} & \textbf{QSC (\%)} & \textbf{TF1 (\%)} & \textbf{MF1 (\%)} \\
  \midrule

  & ValueNet4SPARQL (fine-tuned)
    & \valstd{42.36}{1.18} & \valstd{51.74}{1.43} & \valstd{47.86}{1.09} & \valstd{46.52}{1.26} \\
  \rowcolor{rowgray}
  \cellcolor{white} & Vanilla LLM
    & \valstd{38.91}{1.51} & \valstd{50.28}{1.28} & \valstd{45.37}{1.63} & \valstd{43.96}{1.37} \\
  & CoT-Planner Agent
    & \valstd{46.38}{1.76} & \valstd{58.71}{1.52} & \valstd{52.94}{1.41} & \valstd{51.36}{1.69} \\
  \rowcolor{rowgray}
  \cellcolor{white} & ReAct
    & \valstd{50.27}{1.94} & \valstd{62.48}{1.66} & \valstd{56.72}{1.82} & \valstd{55.08}{1.58} \\
  & AutoGen
    & \valstd{54.86}{1.73} & \valstd{66.35}{2.08} & \valstd{60.47}{1.56} & \valstd{59.21}{1.81} \\
  \rowcolor{oursblue}
  \cellcolor{white}\multirow{-6}{*}{\textit{\spsparql{}}} & \textbf{\textsc{SchemaForge} (Ours)}
    & \valstdb{64.18}{1.29} & \valstdb{74.26}{1.47} & \valstdb{69.83}{1.24} & \valstdb{68.42}{1.53} \\
  \bottomrule
  \end{tabular}
  \end{table*}

\subsection{Graph Allocation on \spsparql{}}
\label{subsec:graph_allocation}

\begin{figure}[t]
\centering
\includegraphics[width=\linewidth]{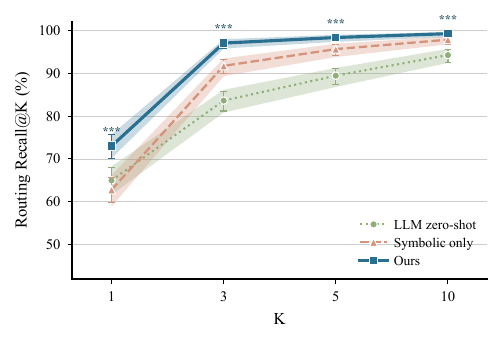}
\caption{Graph allocation Recall@K on \spsparql{}.}
\label{fig:recall_at_k}
\end{figure}

The \spsparql{} results combine two decisions: selecting a graph schema and generating a query under that schema. To isolate the first decision, we evaluate graph allocation using the gold source annotations provided by \spsparql{}. Figure~\ref{fig:recall_at_k} reports routing accuracy across different K.

The zero-shot LLM routing baseline selects the correct graph in $65.0\%$ of examples. \sys{} improves Top-1 accuracy to $73.0\%$, showing that strong schema evidence helps distinguish graphs that are topically related from graphs that can actually support the intended query. When the top three candidate graphs are retained, the gold source is covered in $97.0\%$ of examples. This high Top-3 coverage is important for \sys{} because downstream query generation can operate over a small set of schema-compatible candidates rather than the full graph pool.

Figure~\ref{fig:recall_at_k} also reports a \emph{Symbolic only} ablation that applies the strong-stage schema rerank without weak-retrieval seeding; it underperforms the LLM baseline at $K\!=\!1$ ($62.8\%$ vs $65.0\%$) but exceeds it from $K\!\geq\!3$ ($91.7\%$ vs $83.6\%$), confirming that the two stages are complementary rather than redundant.

\begin{figure}[t]
\centering
\includegraphics[width=\linewidth]{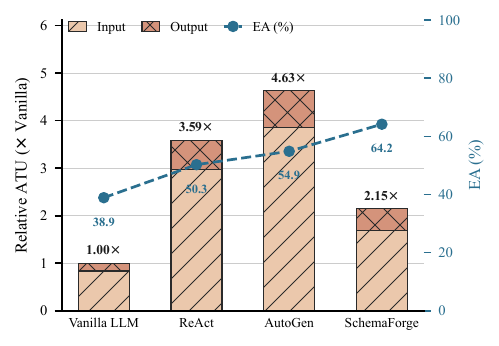}
\caption{Token usage vs.\ EA on \spsparql{}.}
\label{fig:atu_vs_ea}
\end{figure}

\subsection{Ablation Study}
\label{subsec:component_impact}

\begin{table*}[!ht]
\centering
\caption{Component ablation on \spsparql{}.}
\label{tab:ablation}
\setlength{\tabcolsep}{22pt}
\begin{tabular}{lcccc}
\toprule
\textbf{Configuration} & \textbf{EA (\%)} & \textbf{QSC (\%)} & \textbf{TF1 (\%)} & \textbf{MF1 (\%)} \\
\midrule
Ours w/o Decomposer Agent
  & \valstd{39.46}{1.47} & \valstd{55.82}{1.36} & \valstd{45.71}{1.62} & \valstd{44.28}{1.31} \\
\rowcolor{rowgray}
Ours w/o Allocator Agent
  & \valstd{34.72}{1.83} & \valstd{49.64}{1.58} & \valstd{41.83}{1.27} & \valstd{40.16}{1.74} \\
Ours w/o Skeleton Templates
  & \valstd{57.26}{1.58} & \valstd{66.37}{1.74} & \valstd{63.14}{1.46} & \valstd{62.31}{1.63} \\
\rowcolor{rowgray}
Ours w/o Symbolic Validation only
  & \valstd{57.84}{1.36} & \valstd{63.27}{1.94} & \valstd{64.91}{1.22} & \valstd{63.48}{1.57} \\
Ours w/o Counterfactual Check only
  & \valstd{60.53}{1.22} & \valstd{73.18}{1.41} & \valstd{67.52}{1.35} & \valstd{65.91}{1.48} \\
\rowcolor{rowgray}
Ours w/o Verifier Agent (both)
  & \valstd{54.91}{1.69} & \valstd{61.76}{2.08} & \valstd{60.84}{1.44} & \valstd{58.72}{1.63} \\
Ours w/o Agentic Collaborative
  & \valstd{47.68}{1.58} & \valstd{59.43}{1.72} & \valstd{52.36}{1.19} & \valstd{51.27}{1.86} \\
\rowcolor{oursblue}
\textbf{Full \textsc{SchemaForge}}
  & \valstdb{64.18}{1.29} & \valstdb{74.26}{1.47} & \valstdb{69.83}{1.24} & \valstdb{68.42}{1.53} \\
\bottomrule
\end{tabular}
\end{table*}

To assess the contribution of each component, we perform ablations by disabling the \textbf{Decomposer Agent} (no subgoal parsing; pass $q$ directly to the SPARQL generator), the \textbf{Allocator Agent} (replace graph selection with BM25 retrieval over the full KG pool), the \textbf{Skeleton Templates} of the Synthesizer (replace template-guided generation with free-form SPARQL decoding under the same schema slice; allocator and verifier retained), and the \textbf{Verifier Agent} (execute queries without symbolic or counterfactual validation), and by removing \textbf{Agentic Collaboration} (execute all subtasks in a monolithic pipeline without scheduling).

\begin{table}[!ht]
\centering
\caption{Verifier sensitivity to \(\tau_{\mathrm{cf}}\) and \(m\) on \spsparql{}. For the \(\tau_{\mathrm{cf}}\) sweep, \(m=3\) is fixed; for the \(m\) sweep, \(\tau_{\mathrm{cf}}=0.8\) is fixed.}
\label{tab:verifier_sensitivity}
\setlength{\tabcolsep}{8pt}
\begin{tabular}{lccc}
\toprule
\textbf{Setting} & \textbf{EA} & \textbf{Reject/Repair} & \textbf{\# LLM Calls} \\
\midrule
\(\tau_{\mathrm{cf}}=0.6\) & 61.72 & 36.84 & 4.76 \\
\rowcolor{rowgray}
\(\tau_{\mathrm{cf}}=0.7\) & 63.41 & 29.37 & 4.42 \\
\(\tau_{\mathrm{cf}}=0.8\) & \textbf{64.18} & 23.65 & 4.18 \\
\rowcolor{rowgray}
\(\tau_{\mathrm{cf}}=0.9\) & 62.36 & 14.92 & 3.64 \\
\midrule
\(m=1\)        & 61.93 & 18.48 & 3.46 \\
\rowcolor{rowgray}
\(m=3\)        & \textbf{64.18} & 23.65 & 4.18 \\
\(m=5\)        & 64.47 & 25.12 & 5.63 \\
\bottomrule
\end{tabular}
\end{table}

As Table~\ref{tab:ablation} shows, removing the decomposer drops EA by $24.72$ points ($\to39.46$), underscoring the role of compositional parsing for multi-hop queries that span multiple predicates and entity types. Disabling the allocator drops it further to $34.72$ ($-29.46$), the single largest degradation in the table and confirming that schema-aware routing is the dominant axis in the heterogeneous setting where retrieval by surface relevance alone is insufficient. Removing agentic collaboration ($-16.50$ EA) shows that the monolithic baseline still substantially trails the coordinated multi-agent pipeline even when every component is present. Omitting the full verifier costs $9.27$ EA; isolating its two sub-components shows that the symbolic validator and counterfactual check contribute roughly independently ($-6.34$ and $-3.65$ when removed alone), so their combination is close to additive. Notably, removing only the symbolic stage disproportionately hurts QSC ($-10.99$) versus TF1 ($-4.92$), indicating that it primarily filters parse-failure queries while structural correctness is maintained upstream by the decomposer and synthesizer. Replacing skeleton templates with free-form SPARQL decoding (with the allocator and verifier retained) drops EA by $6.92$ points to $57.26$, almost exactly matching the cost of removing the symbolic verifier alone ($-6.34$); templates and symbolic validation thus operate on the same structural-quality axis at different stages of the pipeline (generation-time priors versus post-hoc filtering) and prove roughly substitutable in magnitude. Ranked by EA impact, allocation ($-29.46$) $>$ decomposition ($-24.72$) $>$ agentic collaboration ($-16.50$) $>$ verification ($-9.27$), so schema-aware routing is the most load-bearing component and the verifier acts as a focused safety net rather than the primary driver of accuracy. This ordering also informs deployment trade-offs: under tight budgets, the verifier is the first stage to attenuate while keeping the allocator and decomposer fully active.

The verifier exposes two knobs: the threshold \(\tau_{\mathrm{cf}}\) bounding the answer-set deviation tolerated under counterfactual perturbations, and the perturbation count $m$ per query. Table~\ref{tab:verifier_sensitivity} sweeps both on \spsparql{}.

\emph{(i)} EA traces an inverted-U in \(\tau_{\mathrm{cf}}\), peaking at $64.18\%$ for $0.8$ with a $\sim2$-point drop on either side, so the operating point is not the most permissive setting; over-rejection at $\tau_{\mathrm{cf}}{=}0.6$ is marginally costlier than under-rejection at $0.9$. \emph{(ii)} EA saturates after $m=3$ ($+0.29$ at $m=5$) while \# LLM Calls grows super-linearly, since each rejection triggers a second $m$-perturbation pass; we therefore adopt $m=3$ as Pareto-optimal. \emph{(iii)} EA varies by at most $2.6$ points across $\tau_{\mathrm{cf}}\in[0.6,0.9]$ and $m\in\{1,3,5\}$, so the verifier gains in Table~\ref{tab:ablation} are not threshold-sensitive.

\subsection{Robustness and Efficiency}
\label{subsec:robust_efficiency}

\begin{figure}[t]
\centering
\includegraphics[width=\linewidth]{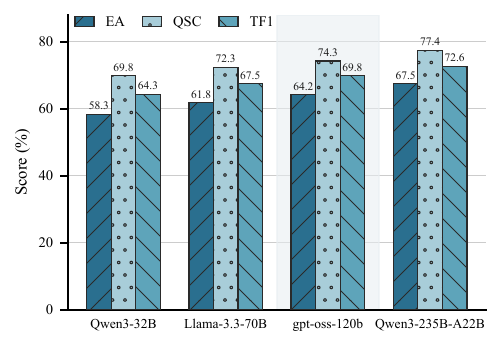}
\caption{Backbone comparison on \spsparql{} (gpt-oss-120b is the main-table backbone).}
\label{fig:backbone_comparison}
\end{figure}

We swap the backbone over four open-weight LLMs spanning a 7$\times$ parameter range: Qwen3-32B~\citep{yang2025qwen3}, Llama-3.3-70B~\citep{grattafiori2024llama3}, gpt-oss-120b~\citep{openai2025gptoss} (main-table backbone), and Qwen3-235B-A22B~\citep{yang2025qwen3} (Figure~\ref{fig:backbone_comparison}). All three metrics improve monotonically with backbone strength (EA $58.3{\to}67.5$; QSC $69.8{\to}77.4$; TF1 $64.3{\to}72.6$) and the $\mathrm{EA}<\mathrm{TF1}<\mathrm{QSC}$ ordering is preserved across all backbones, indicating a consistent structural advantage from schema grounding and verification that is independent of the backbone's intrinsic capability. The $\sim9$-point absolute EA span shows \sys{} is not tied to a frontier model: competitive performance is reachable with smaller open-weight backbones suitable for restricted deployment budgets. The smallest backbone (Qwen3-32B at $58.27$ EA) already outperforms the strongest matched-backbone baseline (AutoGen at $54.86$ EA), so the framework's contribution generalizes beyond the main-table backbone choice.

We further quantify the overhead of modular inference and verification, measured end-to-end across the full agentic pipeline rather than generation alone. Figure~\ref{fig:atu_vs_ea} reports ATU on \spsparql{}, normalized to Vanilla LLM. \sys{} uses $2.15\times$ the tokens of direct prompting but substantially fewer than ReAct ($3.59\times$) and AutoGen ($4.63\times$), while reaching higher EA than both; the overhead is bounded by fixed candidate and verification budgets and amortized through early symbolic rejection, which short-circuits malformed candidates before the more expensive counterfactual stage. The figure places \sys{} above both agent baselines on accuracy-per-token, making the schema-grounded route a Pareto improvement over both general-purpose agent baselines.


\subsection{Case Study}

Appendix~\ref{app:case_studies} presents two case studies. The \spsparql{} success shows the counterfactual verifier catching a load-bearing \texttt{FILTER NOT EXISTS} ($R_{\mathrm{cf}}{=}0.33$); the LC-QuAD~2.0 failure shows a temporal-FILTER literal grounded as a string instead of via \texttt{YEAR(\dots)}, motivating typed literal mutations in the perturbation operator. Together they motivate answer-set perturbation as a check on executable-but-underspecified queries.


\section{Conclusion}
\label{sec:conclusion}

We presented \sys{}, a schema-grounded agentic framework that combines subgoal decomposition, weak-to-strong schema allocation, skeleton-guided synthesis, and counterfactual verification. Across four public benchmarks, \sys{} improves execution accuracy by $+11.50$ EA on macro-average over the strongest matched-backbone baseline, and reaches $73.0\%$/$97.0\%$ Top-1/Top-3 graph allocation on \spsparql{}. Ablations identify schema-aware allocation and counterfactual verification as the dominant contributors, while the skeleton-template and symbolic-validation studies show that structural quality can be enforced at either generation time or post-hoc with comparable effect.

\newpage

\appendix

\section{Notation Table}
\label{app:notation}

\begin{table}[!ht]
\centering
\scriptsize
\setlength{\tabcolsep}{3pt}
\renewcommand{\arraystretch}{1.06}
\caption{Notation used in this paper.}
\label{tab:notation}
\begin{tabular}{@{}p{0.30\columnwidth}p{0.64\columnwidth}@{}}
\toprule
\textbf{Symbol} & \textbf{Meaning} \\
\midrule
\(q\) & Input natural-language question. \\
\(\mathcal{G}=\{G_1,\ldots,G_n\}\) & Collection of \(n\) RDF knowledge graphs. \\
\([n]\) & Graph index set \(\{1,\ldots,n\}\). \\
\(G_j, X_j\) & The \(j\)-th graph and its RDF triples. \\
\(\Sigma_j, \Sigma_{i,j}\) & Schema summary of \(G_j\); schema slice selected for \(g_i\). \\
\(y=(D,\mathbf{a},\hat{\mathbf{s}},\hat{A})\) & Candidate output (decomposition, graph assignments, verified queries, final answer). \\
\(D=(g_1,\ldots,g_k)\) & Decomposition of \(q\) into \(k\) subgoals. \\
\(g_i\) & The \(i\)-th subgoal. \\
\(\mathbf{a}, a_i\) & Graph assignments; candidate graph set for \(g_i\). \\
\(\mathbf{s}, s_{i,j}, \hat{s}_{i,j}\) & Generated query set, query for \((g_i,G_j)\), and selected query. \\
\(I, \hat{A}\) & Answer integration function and final answer. \\
\(\mathrm{Ans}_j(s)\) & Answer set from executing \(s\) on \(G_j\). \\
\midrule
\(\tau_j, e(\cdot)\) & Text serialization of \(G_j\); embedding function. \\
\(S_w, S_s, S_A\) & Weak, strong, and final graph-allocation scores. \\
\(C_i, p\) & Top-\(p\) candidate graphs after weak retrieval. \\
\(\mathrm{PredCov}, \mathrm{TypeCov}, \mathrm{Compat}\) & Predicate coverage, type coverage, and schema compatibility. \\
\(\beta_p,\beta_t,\beta_c,\lambda\) & Allocation weights. \\
\(\mathcal{Q}(g_i,\Sigma_{i,j}), S_Q\) & Schema-constrained query space and query score. \\
\midrule
\(C_{\mathrm{parse}}\) & Indicator for syntactic executability. \\
\(C_{\mathrm{schema}}\) & Indicator for schema compatibility. \\
\(\mathcal{T}(s,\Sigma_{i,j}), T_r, m\) &
Counterfactual perturbation set, \(r\)-th perturbation, and sample count. \\

\(R_{\mathrm{cf}}, \tau_{\mathrm{cf}}\) &
Counterfactual invariance score for non-empty answer sets and its acceptance threshold. \\

\(C_{\mathrm{empty}}\) &
Validation check for empty-answer candidates. \\

\(J(B,C)\) &
Jaccard similarity between non-empty answer sets \(B\) and \(C\). \\
\midrule
\(N, \mathbb{1}[\cdot]\) & Number of evaluation samples and indicator function. \\
\(q_i, q_i^{\mathrm{gold}}\) & Predicted and gold SPARQL queries in evaluation. \\
\(\texttt{compile}(\cdot), \texttt{ans}(\cdot)\) & Query compilation test and execution result. \\
\(\mathrm{EA}, \mathrm{QSC}, \mathrm{TF1}, \mathrm{MF1}\) & Execution accuracy, syntax correctness, triple-level F1, and macro F1. \\
\(P, R, \texttt{TP}_i, \texttt{Pred}_i, \texttt{Gold}_i\) & Precision, recall, true triples, predicted triples, and gold triples. \\
\(\mathrm{ATU}, \texttt{tokens}_{\mathrm{in/out}}\) & Average token usage (input + output) and per-call input/output token counts. \\
\(T, T_{\mathrm{exec}}\) & Per-query timeout and SPARQL execution cost. \\
\bottomrule
\end{tabular}
\end{table}

\section{Implementation Details}
\label{subsec:config}

\subsection{Datasets and Infrastructure}

We evaluate on LC-QuAD~2.0 (6{,}046 test), QALD-9 Plus (136 test), QALD-10 (394 test), and the routable dev subset of \spsparql{} (994 questions over 20 graphs). The \spsparql{} test set is not publicly released; following standard practice, we report on the dev split filtered to the 994 questions answerable by graphs in the 20-graph pool (the remaining questions reference graphs outside the pool). Wikidata benchmarks query the official SPARQL endpoint (\url{https://query.wikidata.org/sparql}); \spsparql{} runs against a local Apache Jena Fuseki, both with a 30\,s per-query timeout. All experiments run on a single Ubuntu~22.04 server with 4$\times$NVIDIA RTX~4090D, 18 vCPUs (AMD EPYC~9754), 128\,GB RAM, and a 500\,GB SSD; software stack is Python~3.12, PyTorch~2.8.0, and CUDA~12.8.

\subsection{Backbones and Hyperparameters}

All agentic methods use the same backbone (gpt-oss-120b for the main table; Qwen3-32B, Llama-3.3-70B, and Qwen3-235B-A22B for the backbone comparison), accessed through official APIs or local deployments with deterministic decoding (temperature~0, $\le 4{,}096$ output tokens). Results are averaged over three random seeds. Main-table runs use $p{=}5$, $r{=}3$, $\lambda{=}0.4$, $(\beta_p,\beta_t,\beta_c){=}(0.5,0.25,0.25)$, $\tau_{\mathrm{cf}}{=}0.8$ with $m{=}3$ perturbations, and $\rho{=}3$. Answers are canonicalized before scoring (deduplication, literal normalization, identifier alignment).

\subsection{Evaluation Metric Details}
\label{app:metrics}

Let \(q_i\) be the predicted query, \(q_i^{\text{gold}}\) the gold query, \(\texttt{ans}(\cdot)\) the execution result, \(N\) the number of evaluation samples, and \(\mathbb{1}[\cdot]\) the indicator function.

\begin{itemize}
    \item \textbf{Execution Accuracy (EA).} Fraction of queries that compile and return the gold answer set:
    \begin{equation*}
        \text{EA} = \tfrac{1}{N}\!\sum_{i=1}^{N}\mathbb{1}\bigl[\texttt{compile}(q_i)\wedge\texttt{ans}(q_i)=\texttt{ans}(q_i^{\text{gold}})\bigr].
    \end{equation*}

    \item \textbf{Query Syntax Correctness (QSC).} Fraction of queries that compile under Apache Jena:
    \begin{equation*}
        \text{QSC} = \tfrac{1}{N}\sum_{i=1}^{N}\mathbb{1}[\texttt{compile}(q_i)].
    \end{equation*}

    \item \textbf{Triple-Level F1 (TF1).} Micro-averaged F1 over WHERE-clause triple patterns (ignoring variable renaming and FILTERs), with per-example correct, predicted, and gold counts \(\texttt{TP}_i,\texttt{Pred}_i,\texttt{Gold}_i\):
    \begin{equation*}
        P = \tfrac{\sum_i |\texttt{TP}_i|}{\sum_i |\texttt{Pred}_i|},\
        R = \tfrac{\sum_i |\texttt{TP}_i|}{\sum_i |\texttt{Gold}_i|},\
        \text{TF1} = \tfrac{2PR}{P+R}.
    \end{equation*}

    \item \textbf{Macro F1 (MF1).} Per-question F1 between the predicted and gold answer sets \(A_i,A_i^{\mathrm{gold}}\) (treated as bags), averaged across questions (QALD protocol):
    \begin{equation*}
        P_i = \tfrac{|A_i\cap A_i^{\mathrm{gold}}|}{|A_i|},\
        R_i = \tfrac{|A_i\cap A_i^{\mathrm{gold}}|}{|A_i^{\mathrm{gold}}|},\
        \text{MF1} = \tfrac{1}{N}\sum_{i=1}^{N}\tfrac{2P_iR_i}{P_i+R_i}.
    \end{equation*}
    \(F1_i\) is taken as \(0\) when a denominator vanishes and as \(1\) when both \(A_i\) and \(A_i^{\mathrm{gold}}\) are empty.

    \item \textbf{Average Token Usage (ATU).} Mean total (input + output) tokens per prediction, summed across all LLM calls in the pipeline:
    \begin{equation*}
        \text{ATU} = \tfrac{1}{N}\sum_{i=1}^{N}\bigl[\texttt{tokens}_{\text{in}}(q_i) + \texttt{tokens}_{\text{out}}(q_i)\bigr].
    \end{equation*}
\end{itemize}


\section{SPARQL Query Skeleton Templates}
\label{app:sparql_templates}


\noindent\textbf{T1. Multi-hop traversal through an intermediate entity.} Joins two predicates via a shared variable; the canonical KGQA pattern for ``X of the Y that has Z'' questions.
\begin{codeblock}
\begin{lstlisting}[style=sparql, basicstyle=\ttfamily\scriptsize]
SELECT DISTINCT ?target ?targetLabel WHERE {
  ?mid wdt:[p1] wd:[anchor] .
  ?mid wdt:[p2] ?target .
  ?target rdfs:label ?targetLabel .
  FILTER(LANG(?targetLabel) = "en")
}
\end{lstlisting}
\end{codeblock}

\noindent\textbf{T2. Typed entities with a numeric filter.} Restricts a class extension by a value constraint on a numeric property (e.g., population, year, area).
\begin{codeblock}
\begin{lstlisting}[style=sparql, basicstyle=\ttfamily\scriptsize]
SELECT DISTINCT ?x ?xLabel WHERE {
  ?x wdt:P31  wd:[class] .
  ?x wdt:[numprop] ?n .
  FILTER(?n [op] [value])
  ?x rdfs:label ?xLabel .
  FILTER(LANG(?xLabel) = "en")
}
\end{lstlisting}
\end{codeblock}

\noindent\textbf{T3. Aggregation with \textcolor{blue!70!black}{\texttt{GROUP BY}} and ranking.} Counts or sums a related set per entity, then orders the results, covering ``which X has the most Y'' questions.
\begin{codeblock}
\begin{lstlisting}[style=sparql, basicstyle=\ttfamily\scriptsize]
SELECT ?g ?gLabel (COUNT(?item) AS ?n) WHERE {
  ?item wdt:[group_prop] ?g .
  ?item wdt:P31 wd:[class] .
  ?g rdfs:label ?gLabel .
  FILTER(LANG(?gLabel) = "en")
}
GROUP BY ?g ?gLabel
ORDER BY DESC(?n)
LIMIT [k]
\end{lstlisting}
\end{codeblock}

\noindent\textbf{T4. Disjunctive pattern via \textcolor{blue!70!black}{\texttt{UNION}}.} Combines alternative property paths, used for ``X that is either Y or Z'' questions and for cross-schema reconciliation on heterogeneous endpoints.
\begin{codeblock}
\begin{lstlisting}[style=sparql, basicstyle=\ttfamily\scriptsize]
SELECT DISTINCT ?person ?personLabel WHERE {
  { wd:[work] wdt:[role1] ?person . }
  UNION
  { wd:[work] wdt:[role2] ?person . }
  ?person rdfs:label ?personLabel .
  FILTER(LANG(?personLabel) = "en")
}
\end{lstlisting}
\end{codeblock}

\noindent{\textit{Note.} These four skeletons illustrate the schema-driven, dataset-agnostic template family used by the Synthesizer; the full library covers single- and multi-hop lookup, numeric and string filtering, aggregation, ranking, and disjunctive patterns with typed slots (\texttt{[entity]}, \texttt{[property]}, \texttt{[class]}, \texttt{[value]}). Wikidata IRIs are shown for illustration; analogous DBpedia and dataset-specific predicates are used when the target endpoint requires them.}

\section{Counterfactual Perturbation Operators}
\label{app:perturbations}

\begin{table}[!ht]
\centering
\small
\setlength{\tabcolsep}{4pt}
\renewcommand{\arraystretch}{1.1}
\caption{Counterfactual perturbation operators used by the Verifier. Each operator targets a single slot type and preserves syntactic validity under $\Sigma_{i,j}$.}
\label{tab:perturbations}
\begin{tabular}{@{}lp{0.62\columnwidth}@{}}
\toprule
\textbf{Operator} & \textbf{Behavior} \\
\midrule
\texttt{PredicateSwap}  & Replace a predicate IRI with another of matching domain/range from $\Sigma_{i,j}$. \\
\texttt{EntitySwap}     & Replace an entity IRI with another sharing at least one \texttt{rdf:type}. \\
\texttt{FilterRelax}    & Drop a \texttt{FILTER} clause. \\
\texttt{FilterTighten}  & Narrow the threshold of a numeric or string \texttt{FILTER}. \\
\texttt{TripleDrop}     & Remove a non-essential triple pattern from the \texttt{WHERE} clause. \\
\bottomrule
\end{tabular}
\end{table}

Table~\ref{tab:perturbations} lists the five type-preserving operators instantiating $\mathcal{T}(s,\Sigma_{i,j})$ (\S\ref{sec:maf}); each replaces one slot (predicate, entity, filter, or triple pattern) while preserving its type and $\operatorname{proj}(s)$, and invalid or trivial variants are skipped from the $R_{\mathrm{cf}}$ expectation. Invariance to a perturbation signals an underspecified slot; the Verifier rejects when the average Jaccard similarity across $m$ perturbations exceeds $\tau_{\mathrm{cf}}$. In practice, predicate- and filter-targeted operators dominate the rejection signal.

\section{Case Studies}
\label{app:case_studies}

\subsection*{Case~1: negation under correct schema allocation}

\noindent\textbf{Dataset:} \textcolor{blue!70!black}{\spsparql{}}, schema \texttt{concert\_singer}. \\
\noindent\textbf{Question:} \emph{Find the names of stadiums that did not host any concert.}

\vspace{2pt}
\noindent\modelbox{Vanilla LLM (gpt-oss-120b) \textcolor{red}{\ding{55}}}
\begin{codeblock}
\begin{lstlisting}[style=sparql, basicstyle=\ttfamily\scriptsize]
SELECT DISTINCT ?name WHERE {
  ?stadium cs:name ?name .
}
\end{lstlisting}
\end{codeblock}
\noindent\textit{Returns all 9 stadiums; the negation requirement is dropped.}

\vspace{2pt}
\noindent\modelbox{\sys{} \textcolor{green!50!black}{\checkmark}}
\begin{codeblock}
\begin{lstlisting}[style=sparql, basicstyle=\ttfamily\scriptsize]
SELECT DISTINCT ?name WHERE {
  ?stadium rdf:type cs:stadium ;
           cs:name  ?name .
  FILTER NOT EXISTS {
    ?concert cs:stadium_id ?stadium .
  }
}
\end{lstlisting}
\end{codeblock}

\noindent\textit{What worked.} The Allocator routed to \texttt{concert\_singer} (\(S_A=0.86\); runner-up \(\le 0.42\)), the Decomposer split into \emph{all stadiums} and \emph{stadiums hosting a concert}, and the Synthesizer combined them via a negation skeleton (\texttt{FILTER NOT EXISTS} over the typed-entity base of T2). The Verifier's perturbation that dropped \texttt{FILTER NOT EXISTS} returned 9 vs.\ 3 stadiums (\(R_{\mathrm{cf}}=0.33\)), confirming the negation is load-bearing.

\subsection*{Case~2: mishandled temporal filter}

\noindent\textbf{Dataset:} \textcolor{blue!70!black}{LC-QuAD~2.0}, Wikidata target. \\
\noindent\textbf{Question:} \emph{List all the films directed by Akira Kurosawa after 1960.}

\vspace{2pt}
\noindent\textbf{Gold SPARQL.}
\begin{codeblock}
\begin{lstlisting}[style=sparql, basicstyle=\ttfamily\scriptsize]
SELECT DISTINCT ?film WHERE {
  ?film wdt:P57  wd:Q8006 .          # director: Akira Kurosawa
  ?film wdt:P577 ?date .             # publication date
  FILTER(YEAR(?date) > 1960)
}
\end{lstlisting}
\end{codeblock}

\noindent\modelbox{\sys{} \textcolor{red}{\ding{55}}}
\begin{codeblock}
\begin{lstlisting}[style=sparql, basicstyle=\ttfamily\scriptsize]
SELECT DISTINCT ?film WHERE {
  ?film wdt:P57  wd:Q8006 .
  ?film wdt:P577 ?date .
  FILTER(?date > "1960")
}
\end{lstlisting}
\end{codeblock}

\noindent\textit{What failed.} Allocation and decomposition succeeded; the Synthesizer instantiated T2 but bound the \texttt{FILTER} right-hand side to the string \texttt{"1960"} instead of \texttt{YEAR(?date)}, yielding a lexicographic comparison over \texttt{xsd:dateTime} that preserves most post-1960 films. Since the answer is non-empty and the Verifier's perturbations target predicates not FILTER literals, the query is accepted (high TF1, failing EA). This motivates typed literal mutations (\texttt{YEAR}, \texttt{xsd:integer}) in both perturbation and skeleton libraries.

\clearpage
\newpage
\twocolumn



\bibliographystyle{ACM-Reference-Format}
\bibliography{CIKM/acmart}

\end{document}